\documentclass[sigconf,10pt]{acmart}

\renewcommand\footnotetextcopyrightpermission[1]{}
\usepackage{enumitem}
\usepackage{microtype} 
\usepackage[ruled,linesnumbered]{algorithm2e}
\usepackage{amsmath}
\usepackage{multirow}
\usepackage{booktabs}
\usepackage{subfigure}
\usepackage{xcolor}
\usepackage{pifont}
\usepackage{fp}
\usepackage{caption}
\newlength\barwd
\newcommand\tbar[4][blue!15]{%
  \FPeval\result{round((#3)/#2:4)}%
  \rlap{\textcolor{#1}{\hspace*{\dimexpr-\tabcolsep+.5\arrayrulewidth}%
        \rule[-.05\ht\strutbox]{\result\barwd}{.95\ht\strutbox}}}%
  \makebox[\barwd][r]{#4}}

\begin{document}

\title{PowerLens: Taming LLM Agents for Safe and Personalized Mobile Power Management}

\author{Xingyu Feng$^{1}$, Chang Sun$^{1}$, Yuzhu Wang$^{1}$, Zhangbing Zhou$^{1}$, Chengwen Luo$^{2}$, Zhuangzhuang Chen$^{3}$, Xiaomin Ouyang$^{3}$, Huanqi Yang$^{4}$}
\affiliation{%
\institution{$^1$China University of Geosciences (Beijing), $^2$Shenzhen University, $^3$Hong Kong University of Science and Technology, $^4$City University of Hong Kong}
\country{}}
\renewcommand{\shortauthors}{Feng et al.}

\begin{abstract}
Battery life remains a critical challenge for mobile devices, yet existing power management mechanisms rely on static rules or coarse-grained heuristics that ignore user activities and personal preferences.
We present PowerLens, a system that tames the reasoning power of Large Language Models (LLMs) for safe and personalized mobile power management on Android devices.
The key idea is that LLMs' commonsense reasoning can bridge the semantic gap between user activities and system parameters, enabling zero-shot, context-aware policy generation that adapts to individual preferences through implicit feedback.
PowerLens employs a multi-agent architecture that recognizes user context from UI semantics and generates holistic power policies across 18 device parameters.
A PDL-based constraint framework verifies every action before execution, while a two-tier memory system learns individualized preferences from implicit user overrides through confidence-based distillation, requiring no explicit configuration and converging within 3--5 days.
Extensive experiments on a rooted Android device show that PowerLens achieves {81.7\%} action accuracy and {38.8\%} energy saving over stock Android, outperforming rule-based and LLM-based baselines, with high user satisfaction, fast preference convergence, and strong safety guarantees, with the system itself consuming only {0.5\%} of daily battery capacity.
\end{abstract}

\maketitle

\section{Introduction}

 \begin{figure}[tbp]
\centering
\includegraphics[width=\columnwidth]
{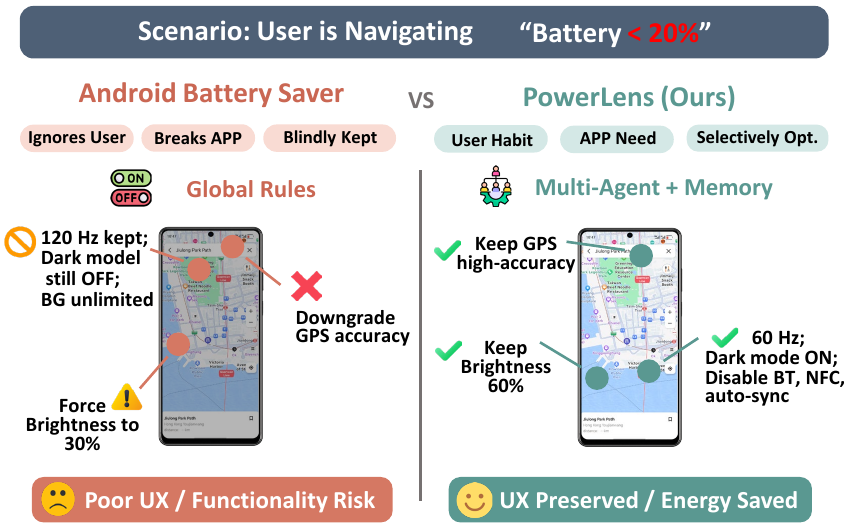}
\caption{\textbf{Traditional power saver vs.\ PowerLens.} Global rules degrade navigation by throttling GPS and dimming brightness; PowerLens preserves critical resources and learned preferences.}
\label{fig:motivation}
\vspace*{-0.1in}
\end{figure}

Battery life remains one of the most persistent pain points for smartphone users.
Despite decades of hardware and OS-level optimization, mobile devices still struggle to last through a full day of active use~\cite{gupta20243,chen2015smartphone,balasubramanian2009energy}.
As smartphones become the primary computing platform for billions of users, supporting tasks from navigation and video streaming to mobile gaming and video conferencing, their power management systems face an increasingly complex optimization landscape.
Dozens of adjustable parameters, including screen brightness, refresh rate, CPU governor, location mode, and connectivity radios, must be jointly balanced against diverse and context-dependent user requirements.

Existing approaches to mobile power management can be broadly classified into three categories: hardware-level, OS-level, and learning-based techniques.
Hardware-level approaches such as DVFS \cite{lin2023workload, kim2022ztt} reactively adjust CPU frequency based on instantaneous utilization using governors like \texttt{ondemand} and \texttt{schedutil}.
While effective at the CPU level, they are oblivious to the broader device context: the same CPU load during navigation versus a background sync has fundamentally different optimization opportunities.
OS-level mechanisms such as Android's Adaptive Battery and App Standby Buckets~\cite{android_adaptive_battery} apply coarse-grained heuristics based on app usage recency to classify apps into priority tiers, which ignore both app-specific functional requirements (e.g., navigation requires high-accuracy GPS) and individual user preferences (e.g., a user who always prefers high brightness while reading).
Orthogonally, learning-based approaches~\cite{nawrocki2020adaptive,sunder2025smartapm} attempt to learn optimal policies from historical usage traces and system telemetry, but typically require extensive training data, struggle to generalize across devices and usage patterns, and operate on low-level numerical signals without semantic understanding of user activities.

\begin{figure*}
\centering
\subfigure[Resource levels vary by activity.]{
\begin{minipage} {0.24\linewidth}
\centering
\includegraphics[width=0.99\linewidth]{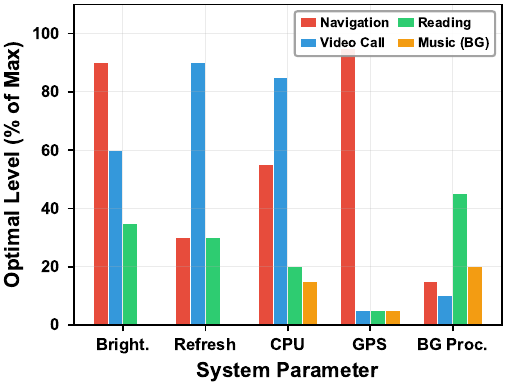}
\label{fig:challenge_params}
\vspace{-4mm}
\end{minipage}%
}
\hfill
\subfigure[Context-blind wastes 19--50\%.]{
\begin{minipage} {0.24\linewidth}
\centering
\includegraphics[width=0.99\linewidth]{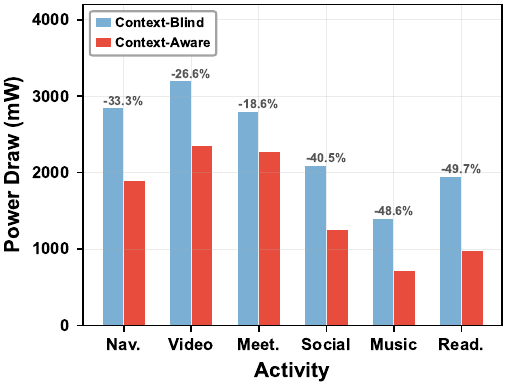}
\label{fig:challenge_waste}
\vspace{-4mm}
\end{minipage}%
}
\hfill
\subfigure[Override patterns vary by user.]{
\begin{minipage} {0.24\linewidth}
\centering
\includegraphics[width=0.99\linewidth]{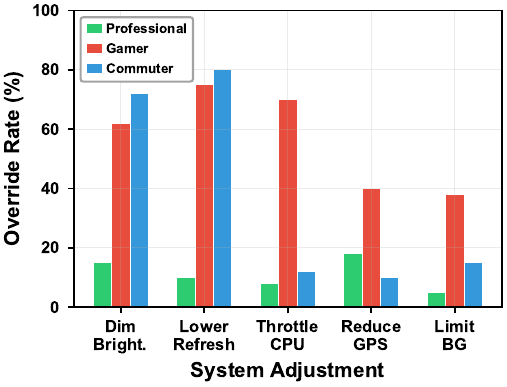}
\label{fig:challenge_override}
\vspace{-4mm}
\end{minipage}%
}
\hfill
\subfigure[LLM violation types across models.]{
\begin{minipage} {0.24\linewidth}
\centering
\includegraphics[width=0.99\linewidth]{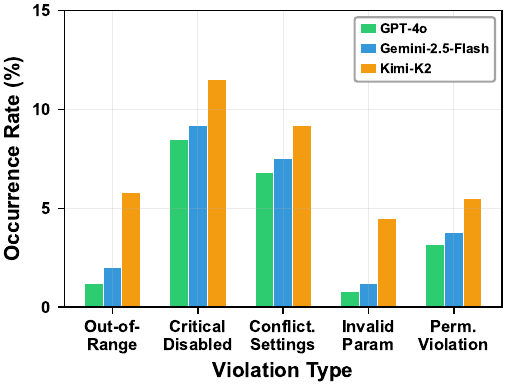}
\label{fig:challenge_safety}
\vspace{-4mm}
\end{minipage}%
}
\vspace{-2mm}
\caption{\textbf{Preliminary studies on challenges of mobile power management.} }
\label{fig:challenges}
\vspace{-0.1in}
\end{figure*}

In this work, we introduce PowerLens, a Large Language Model (LLM)-powered personalized power management system for mobile devices.
The key insight is that LLMs can serve as zero-shot system-level reasoners, bridging the semantic gap between user activities and device parameters. For example, an LLM can understand that ``the user is navigating to an unfamiliar destination'' implies GPS accuracy and screen visibility are critical, while background sync and high refresh rates can be safely reduced.
PowerLens combines this reasoning capability with domain-specific knowledge about device constraints and a personalized memory system that learns user preferences through implicit feedback (Fig.~\ref{fig:motivation}).

Unlike existing LLM-based mobile agents~\cite{lee2024mobilegpt,wen2024autodroid,zhang2025appagent} that focus on UI-level task automation (e.g., ``send a message'' or ``book a restaurant''), PowerLens operates at the \textit{system-level resource management} layer, adjusting hardware parameters that are invisible to the user interface but critically affect both battery life and user experience.
This introduces three unique challenges:

\begin{enumerate}[nosep,leftmargin=*]

\item \textbf{Context-Aware Policy Generation.}
Effective power management requires jointly reasoning about what the user is doing, what the current app needs, and what the device supports.
As shown in Fig.~\ref{fig:challenge_params}, optimal resource levels vary dramatically across activities (navigation demands high GPS and brightness while music playback needs almost none), and applying context-blind defaults wastes 19--50\% power compared to a context-aware oracle (Fig.~\ref{fig:challenge_waste}).
A single app (e.g., a browser) may exhibit vastly different resource requirements depending on the content (video playback vs.\ text reading), and different users may have different tolerance for performance degradation.
\textbf{Our approach:} We address this by designing a multi-agent architecture where specialized agents decompose the problem: the Activity Agent infers semantic context from UI states, and the Policy Agent synthesizes a holistic strategy informed by device capabilities and user preferences.

\item \textbf{Personalized Preference Learning from Implicit Feedback.}
Users rarely provide explicit power management preferences (e.g., ``I prefer 60\,Hz for reading'').
Instead, their preferences are revealed implicitly through \textit{manual overrides}: when a user increases brightness after the system dimmed it, this signals dissatisfaction.
Fig.~\ref{fig:challenge_override} illustrates that override patterns are highly user-specific: a Gamer frequently rejects CPU and refresh-rate throttling, while a Commuter overrides display dimming but accepts compute savings.
\textbf{Our approach:} We design a two-tier memory system with a state differencing mechanism that detects such overrides and a confidence-based distillation process that gradually promotes observed patterns into stable preference rules.

\item \textbf{Safe and Verifiable Execution.}
LLMs can hallucinate invalid parameter values or propose changes that would break app functionality (e.g., disabling GPS during navigation).
Fig.~\ref{fig:challenge_safety} shows that raw LLM outputs produce over 20\% problematic actions even from frontier models.
\textbf{Our approach:} We introduce a constraint verification framework based on Propositional Dynamic Logic (PDL) that validates every LLM-generated action against device-specific capabilities and app-category safety invariants before execution, ensuring that creative optimization strategies never violate critical constraints.
\end{enumerate}

We implement PowerLens \footnote{Project page with demo videos: \url{https://anonymous-powerlens.github.io/}} as an Android system application requiring root access and evaluate it on PowerLensBench, a new benchmark we constructed, spanning mainstream apps across 7 usage categories.
Compared to stock Android and three baselines (Battery Saver, Rule-Based, Single-Agent LLM), PowerLens achieves the best balance of energy saving ({38.8\%}) and user experience ({4.3}/5.0, {81.7\%} action accuracy) while maintaining the lowest safety violation rate ({0.6\%}).
The two-tier memory system converges within {3--5 days}, and the PDL checker eliminates {96.5\%} of raw LLM-generated safety violations.

Our main contributions are:
\begin{itemize}[nosep,leftmargin=*]
\item PowerLens is the first system to employ LLM agents for personalized, context-aware power management on mobile devices, decomposing the problem via a multi-agent architecture into activity recognition, policy generation, execution verification, and implicit feedback collection.
\item A two-tier memory system with implicit feedback detection via state differencing and confidence-based preference distillation, enabling personalized preference learning without explicit user configuration and converging within {3--5 days} across diverse user profiles.
\item A PDL-based constraint verification framework that ensures LLM-generated power policies respect device capabilities and app-specific safety invariants, reducing violations by {96.5\%} while preserving LLM reasoning flexibility.
\item Extensive experiments across 48 tasks, 7 app categories, and diverse user profiles on real hardware, along with an in-the-wild user study, demonstrating {38.8\%} energy saving with {81.7\%} action accuracy and {4.3}/5.0 user experience.
\end{itemize}

\begin{table*}
\centering
\caption{\textbf{Parameters controlled by PowerLens.} 18 parameters across 5 categories, spanning a joint state space of {$>$}\,$10^{17}$ configurations.}
\label{tab:power_params}
\fontsize{5}{6}\selectfont
\resizebox{\textwidth}{!}{%
\begin{tabular}{lrllllllccrc}
\toprule
\textbf{Category} & \textbf{Power\textsuperscript{\P}} & \textbf{Parameter} & \textbf{HW Component} & \textbf{Range} & \textbf{Default\textsuperscript{\dag}} & \textbf{Control API} & \textbf{Granularity} & \textbf{Root} & \textbf{Latency} & \textbf{$|\mathcal{S}|$} & \textbf{Impact\textsuperscript{\ddag}} \\
\midrule
\multirow{5}{*}{Display} & \multirow{5}{*}{42\%}
& Brightness        & OLED panel              & 0--4096                        & Auto      & Settings.System   & Continuous  & ---           & {<}\,1\,s    & 4,097  & \textbf{\textcolor{blue}{H}} \\
&& Refresh rate      & Display controller      & \{30, 60, 90, 120\}\,Hz        & 120\,Hz   & Settings.System   & Discrete    & ---           & {<}\,1\,s    & 4      & \textbf{\textcolor{blue}{H}} \\
&& Screen timeout    & Display controller      & 15\,s--30\,min                 & 30\,s     & Settings.System   & Discrete    & ---           & Deferred     & 7      & \textbf{\textcolor{cyan}{M}} \\
&& Dark mode         & OLED panel              & On/Off                          & Off       & UiModeManager     & Binary      & ---           & {<}\,1\,s    & 2      & \textbf{\textcolor{gray}{L}} \\
&& Auto-rotation     & Gyroscope/Accel.        & On/Off                          & On        & Settings.System   & Binary      & ---           & {<}\,1\,s    & 2      & \textbf{\textcolor{gray}{L}} \\
\midrule
\multirow{5}{*}{Connectivity} & \multirow{5}{*}{25\%}
& Wi-Fi             & Wi-Fi radio (WCN7851)   & On/Off                          & On        & Settings.Global   & Binary      & ---           & 1--3\,s      & 2      & \textbf{\textcolor{cyan}{M}} \\
&& Bluetooth         & BT radio (WCN7851)      & On/Off                          & On        & Settings.Global   & Binary      & ---           & 1--2\,s      & 2      & \textbf{\textcolor{gray}{L}} \\
&& NFC               & NFC controller          & On/Off                          & On        & Settings.Global   & Binary      & ---           & {<}\,1\,s    & 2      & \textbf{\textcolor{gray}{L}} \\
&& Mobile data       & 5G/LTE modem            & On/Off                          & On        & TelephonyManager  & Binary      & $\checkmark$  & 2--5\,s      & 2      & \textbf{\textcolor{cyan}{M}} \\
&& Location mode     & GNSS receiver           & Off/Dev./Saving/High            & High      & Settings.Secure   & Discrete    & ---           & 1--3\,s      & 4      & \textbf{\textcolor{blue}{H}} \\
\midrule
\multirow{3}{*}{Compute} & \multirow{3}{*}{20\%}
& CPU governor      & SoC (Snapdragon 8G3)    & powersave/schedutil/perf.       & schedutil & sysfs             & Discrete    & $\checkmark$  & {<}\,10\,ms  & 3      & \textbf{\textcolor{blue}{H}} \\
&& CPU cores online  & SoC (per-core)          & 8-bit binary mask               & 0xFF      & sysfs             & Bitmask     & $\checkmark$  & {<}\,10\,ms  & 256    & \textbf{\textcolor{blue}{H}} \\
&& BG process limit  & SoC / RAM               & \{$-$1, 0, 1, 2, 3, 4\}        & $-$1      & Settings.Global   & Discrete    & ---           & Deferred     & 6      & \textbf{\textcolor{cyan}{M}} \\
\midrule
\multirow{2}{*}{Audio} & \multirow{2}{*}{5\%}
& Media volume              & Speaker / DAC   & 0--160  & ---  & AudioManager      & Continuous  & ---           & {<}\,1\,s    & 161    & \textbf{\textcolor{gray}{L}} \\
&& Notif./Ring/Alarm vol.\textsuperscript{$\ast$}    & Speaker / DAC   & 0--16 ea.   & ---  & AudioManager  & Discrete    & ---           & {<}\,1\,s    & 17 each & \textbf{\textcolor{gray}{L}} \\
\midrule
Sync & 8\% & Auto-sync       & CPU / Radio             & On/Off                          & On        & ContentResolver   & Binary      & ---           & Async        & 2      & \textbf{\textcolor{cyan}{M}} \\
\bottomrule
\end{tabular}%
}
\vspace{2pt}
\parbox{\textwidth}{\raggedright\scriptsize
\textsuperscript{\dag}Default on test device (OnePlus ACE 5, Snapdragon 8 Gen 3, Android 15 with KernelSU). \quad
\textsuperscript{\P}Representative category-level power proportions under typical mixed use~\cite{gupta20243}. \quad
\textsuperscript{\ddag}Impact: \textbf{\textcolor{blue}{H}}igh ({>}15\% category saving), \textbf{\textcolor{cyan}{M}}edium (5--15\%), \textbf{\textcolor{gray}{L}}ow ({<}5\%). \quad
$|\mathcal{S}|$: discrete states per parameter; $\prod|\mathcal{S}_i| > 10^{17}$. \textsuperscript{$\ast$}Notif./Ring/Alarm: three independent streams, 17 states each.
}
\vspace*{-0.15in}
\end{table*}

\section{Background and Motivation}

\subsection{Mobile Power Management Landscape}

\begin{figure}[tbp]
\centering
\includegraphics[scale=0.33]{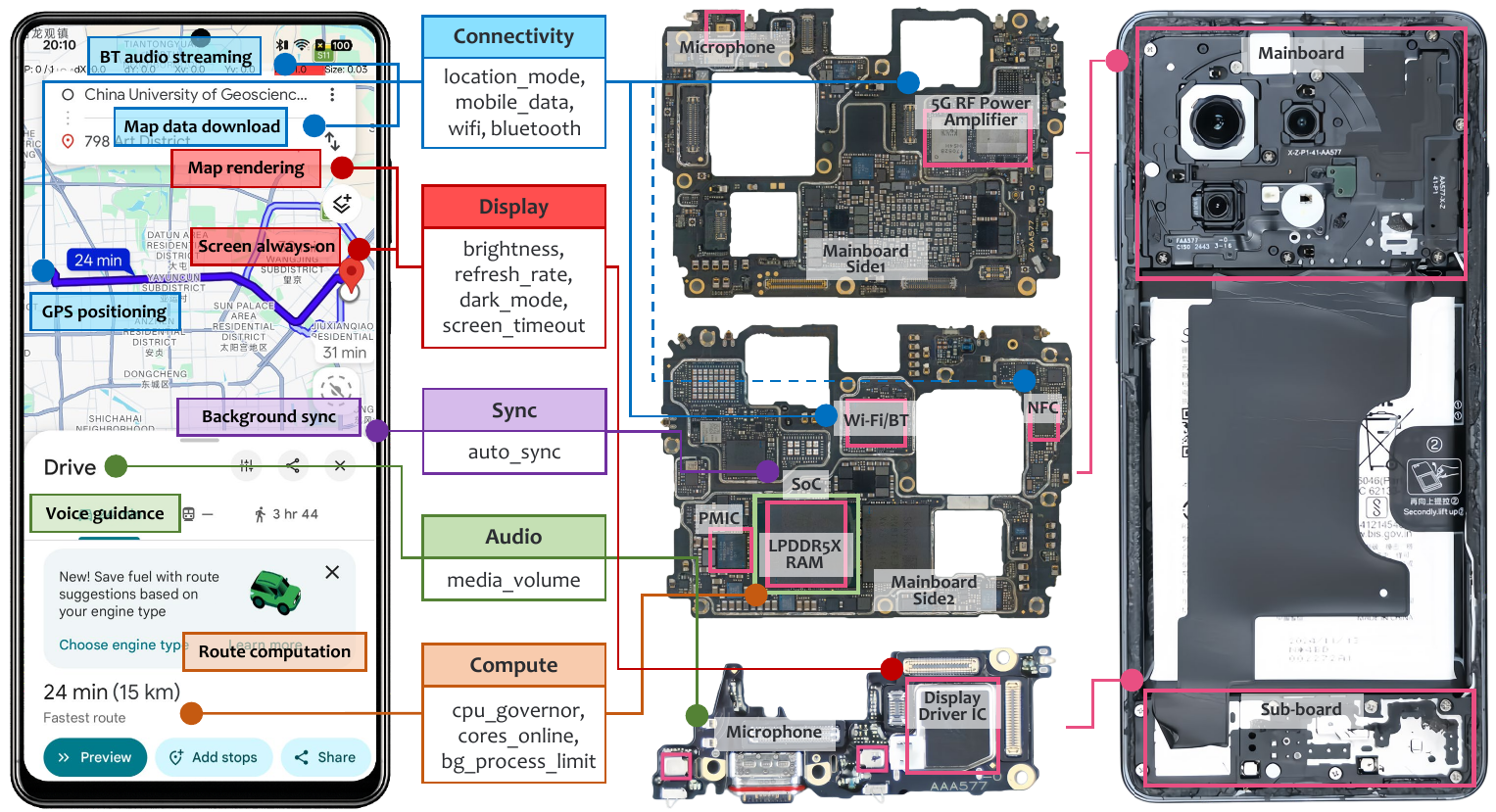}
\caption{\textbf{Hardware activated during navigation.} A single app engages components across the mainboard and daughter board, spanning all five parameter categories in Table~\ref{tab:power_params}.}
\label{fig:hardware}
\vspace*{-0.25in}
\end{figure}

Table~\ref{tab:power_params} summarizes the 18 parameters that PowerLens controls, spanning five categories:
\textit{display} (brightness, refresh rate, timeout, dark mode, auto-rotation), \textit{connectivity} (Wi-Fi, Bluetooth, NFC, mobile data, location mode), \textit{compute} (CPU governor, core mask, background process limit), \textit{audio} (media, notification, ringtone, and alarm volumes), and \textit{sync} (auto-sync).
Even the 18 controlled parameters span a combinatorial action space exceeding $10^{17}$ valid configurations.
Fig.~\ref{fig:hardware} illustrates the hardware complexity behind a single navigation session: turn-by-turn guidance simultaneously draws power from the SoC (route computation), GNSS receiver (positioning), 5G modem and Wi-Fi radio (map downloads), OLED panel (high-brightness map rendering), speaker (voice prompts), and Bluetooth radio (hands-free audio), engaging components distributed across both sides of the mainboard and the daughter board.
Saving power in such a scenario requires coordinated trade-offs across all five parameter categories (13 of 18 parameters active), a task beyond any single-component governor.

Current mechanisms operate at different granularities.
At the lowest level, DVFS governors such as \texttt{schedutil} dynamically scale CPU frequency based on load, while advanced governors like GearDVFS~\cite{lin2023workload} account for concurrent workload interference.
At the OS level, Android Adaptive Battery classifies apps into standby buckets (Active, Working Set, Frequent, Rare, Restricted), applying progressively stricter scheduling and network limitations.
Doze mode aggressively restricts background activity when idle.

These mechanisms share a fundamental limitation: \textit{they lack semantic understanding of user activities}.
Adaptive Battery treats a food delivery app in the same standby bucket regardless of whether the user is actively tracking an order or has not opened it in days.
DVFS governors cannot distinguish CPU load from a video call (where frame drops degrade experience) versus background indexing (where throttling is imperceptible), leaving significant optimization opportunities unexploited.

\subsection{LLMs as System-Level Reasoners}

The recent success of LLM-based agents in mobile task automation demonstrates that LLMs can effectively reason about complex software systems:
AutoDroid~\cite{wen2024autodroid,wen2025autodroid} plans multi-step actions without training; MobileGPT~\cite{lee2024mobilegpt} augments LLMs with task memory for recurring tasks; AutoIOT~\cite{shen2025autoiot} bridges natural language intent and low-level device operations.

These results suggest LLMs possess three capabilities essential for power management:
(1)~\textit{semantic understanding}, inferring that ``Google Maps showing turn-by-turn directions'' implies the user needs GPS accuracy;
(2)~\textit{commonsense reasoning}, knowing that reducing brightness during nighttime reading is acceptable but during outdoor navigation is not; and
(3)~\textit{compositional planning}, generating coordinated policies across multiple parameters (e.g., simultaneously reducing refresh rate \textit{and} switching to a power-saving CPU governor for a reading session).
However, system-level resource management introduces challenges absent from task automation: actions are continuous rather than discrete UI clicks, consequences are latent, and safety constraints are strict (a wrong GPS setting can cause navigation failure).

\subsection{Motivating Scenarios}

Three scenarios illustrate these limitations:

\noindent\textbf{Scenario 1: Context-blind optimization.}
A user opens a news-reading app. The stock system maintains 120\,Hz refresh rate and high CPU frequency, wasting power on resources that reading does not need.
An LLM-powered system recognizes the reading activity and reduces the refresh rate to 60\,Hz and switches to a power-saving CPU governor.

\noindent\textbf{Scenario 2: Ignoring personal preferences.}
Two users open the same video streaming app at night: User A prefers low brightness (dark room), User B prefers medium brightness (public transit).
Stock Android applies the same adaptive brightness to both, satisfying neither.
PowerLens learns these distinct preferences through implicit feedback: when User B repeatedly overrides the system's dimming, the memory system captures this pattern and stops overriding in future sessions.

\noindent\textbf{Scenario 3: Safety-unaware power saving.}
When battery drops below 20\% during navigation, Android's Low Power Mode force-dims the screen and throttles GPS (Fig.~\ref{fig:motivation}, left), breaking the navigation experience.
PowerLens recognizes that GPS and screen visibility are critical for navigation, preserves them, and saves power through other parameters instead (e.g., reducing refresh rate) (Fig.~\ref{fig:motivation}, right).


\begin{figure}
\centering
\includegraphics[scale=0.43]{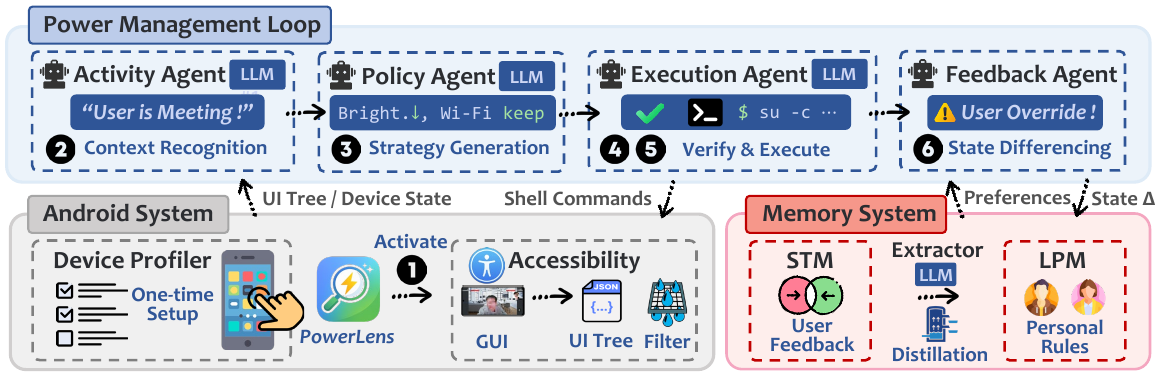}
\caption{\textbf{PowerLens system overview.} Each cycle: \ding{202}~Accessibility captures UI tree, \ding{203}~Activity Agent recognizes context, \ding{204}~Policy Agent generates strategy, \ding{205}\ding{206}~Execution Agent verifies and applies via shell commands, \ding{207}~Feedback Agent detects user overrides. The Memory System stores preferences for personalization.}
\label{fig:overview}
\vspace*{-0.15in}
\end{figure}

\section{System Design}
\label{sec:design}

As shown in Fig.~\ref{fig:overview}, PowerLens comprises two tightly coupled subsystems: the \textit{Android System} layer and the \textit{Power Management Loop}.
The Android System layer provides a \textbf{Device Profiler} that enumerates adjustable parameters and their valid ranges (the \textit{action space library}), and an \textbf{Accessibility} framework that converts the foreground app's GUI into structured UI XML for semantic activity recognition.
The Power Management Loop orchestrates four agents (Activity, Policy, Execution, and Feedback) in a closed-loop control system, connected by a two-tier \textbf{Memory System} whose asynchronous Extractor performs LLM-based intent analysis during device idle time.
In total, the system involves five LLM invocation points: four synchronous calls per decision cycle (Activity, Policy, and two Execution stages) plus one asynchronous call in the Extractor for preference distillation.
Unlike traditional approaches~\cite{sunder2025smartapm,nawrocki2020adaptive}, PowerLens leverages LLM reasoning for zero-shot, personalized policy generation with safety verification.

\subsection{Multi-Agent Power Management Loop}
\label{sec:agents}

PowerLens decomposes the power management problem into four specialized agents, Activity ($M_a$), Policy ($M_p$), Execution ($M_e$), and Feedback ($M_f$), that form a sequential pipeline with a feedback loop.
Of these, Activity, Policy, and Execution invoke the LLM; the Feedback Agent uses deterministic state differencing without LLM inference.
This modular design enables independent testing and updating of each agent, graceful degradation via fallback policies, and interpretable reasoning.

\subsubsection{Activity Agent.}
\label{sec:activity}
The Activity Agent $M_a$ is responsible for transforming raw device signals into a semantically rich context representation.
It takes three inputs: (1)~the \textit{UI tree}, obtained via Android's Accessibility framework, which captures the current screen's widget hierarchy as a structured JSON document; (2)~the \textit{device context} $\sigma$, a snapshot of the device's current state including battery level, connectivity status, sensor readings, and all adjustable parameter values; and (3)~the \textit{recent app history}, which records the apps used within the past hour along with their usage durations.

Using these inputs, $M_a$ performs three functions.
\textit{Activity recognition}: the UI tree and recent app history are fed to an LLM that identifies (i)~the user's current high-level \textit{activity type} corresponding to the app category (e.g., \textsc{Navigation}, \textsc{Video\_Watching}, \textsc{Gaming}, \textsc{Reading}), and (ii)~a finer-grained \textit{sub-activity} describing the specific interaction mode within the app (e.g., ``actively watching'' vs.\ ``browsing comments'' in a video app), along with a recognition certainty score and critical level (high/medium/low) indicating how interruption-sensitive the activity is.
\textit{Context extraction}: the agent extracts structured context tags including the foreground app, UI state, inferred user intent, and environmental cues (e.g., ``user is in a meeting room based on connected Wi-Fi SSID'').
\textit{Context signature construction}: the extracted features are discretized into a \textit{context signature} $\mathbf{s}$, a compact tuple of the factors that most influence power policy selection:
\begin{equation}
\label{eq:context}
\mathbf{s} = \big(\mathit{cat}(A_t),\; \mathit{type}(A_t),\; \mathit{bucket}(B_t),\; \mathit{period}(T_t)\big),
\end{equation}
where $\mathit{cat}(A_t)$ is the app category, $\mathit{type}(A_t)$ the recognized sub-activity, $\mathit{bucket}(B_t) \in \{\textsc{High}, \textsc{Mid}, \textsc{Low}\}$ discretizes battery level, and $\mathit{period}(T_t)$ captures temporal context.
This compact discrete signature enables efficient LPM retrieval despite the continuous nature of the underlying state space.

\subsubsection{Policy Agent.}
\label{sec:policy}
$M_p$ receives the device context $\sigma$, activity recognition result, memory-derived preferences (Sec.~\ref{sec:memory}), PDL safety constraints $\Phi$ (Sec.~\ref{sec:constraints}), and device capabilities, and generates a structured policy $\pi$ through \textit{priority arbitration} across three memory sources:
\begin{equation}
\pi = \textsc{Arbitrate}\big(\underbrace{C_{\text{STM}}}_{\text{highest}},\; \underbrace{R_{\text{LPM}}(\mathbf{s})}_{\text{medium}},\; \underbrace{G_{\text{LPM}}}_{\text{lowest}}\big),
\end{equation}
where $C_{\text{STM}}$ denotes active constraints from STM (user manual overrides that must be obeyed unconditionally), $R_{\text{LPM}}(\mathbf{s})$ denotes context-specific rules retrieved from LPM via the context signature, and $G_{\text{LPM}}$ is the general user profile used as a fallback when no context rule matches.
The memory arbitration follows a strict priority: \textit{STM user locks} $>$ \textit{LPM context rules} $>$ \textit{LPM general profile}.
Safety constraints $\Phi$ are enforced orthogonally: $M_p$ is instructed to respect them during generation, and the Execution Agent independently verifies compliance before execution (Sec.~\ref{sec:constraints}), providing a two-stage defense.
The output policy is a set of actions:
\begin{align}
\pi &= \{a_1, a_2, \ldots, a_m\}, \notag \\
a_i &= (\mathit{target}_i, \mathit{action}_i, \mathit{value}_i, \mathit{priority}_i, \mathit{reason}_i),
\end{align}
where each action specifies a target parameter, control action (\textsc{Keep}, \textsc{Set}, \textsc{Enable}, \textsc{Disable}, \textsc{Lock}, \textsc{Defer}), target value, priority level, and a natural language reason for logging and downstream intent analysis.

\subsubsection{Execution Agent.}
The Execution Agent $M_e$ serves as the bridge between LLM-generated strategies and the physical device.
It performs three sequential functions via \textit{two separate LLM calls}: verification and command generation are deliberately separated so that the generation call receives only approved actions, preventing hallucinated verification reasoning from leaking into executable commands.

\textit{Legality verification} (LLM call 1): each proposed action is checked by the LLM against the device capabilities profile to ensure the target value falls within the valid range (e.g., brightness value within $[0, 4096]$, refresh rate is one of $\{30, 60, 90, 120\}$) and against PDL constraints to ensure no safety invariant is violated (Sec.~\ref{sec:constraints}).
Actions that fail verification are either corrected to the nearest valid value or rejected, with a fallback to a conservative default.
Only approved actions proceed to the next stage.

\textit{Shell command generation} (LLM call 2): approved actions are translated by the LLM into executable Android shell commands via the \texttt{su} root shell (e.g., setting refresh rate requires three synchronized \texttt{settings put} commands).

\textit{State synchronization}: after execution, the agent reads back the actual system state to confirm that commands took effect, and updates the STM's \textit{last known state} snapshot for the Feedback Agent's subsequent state differencing.

 \begin{figure}[tbp]
\centering
\includegraphics[width=\columnwidth]{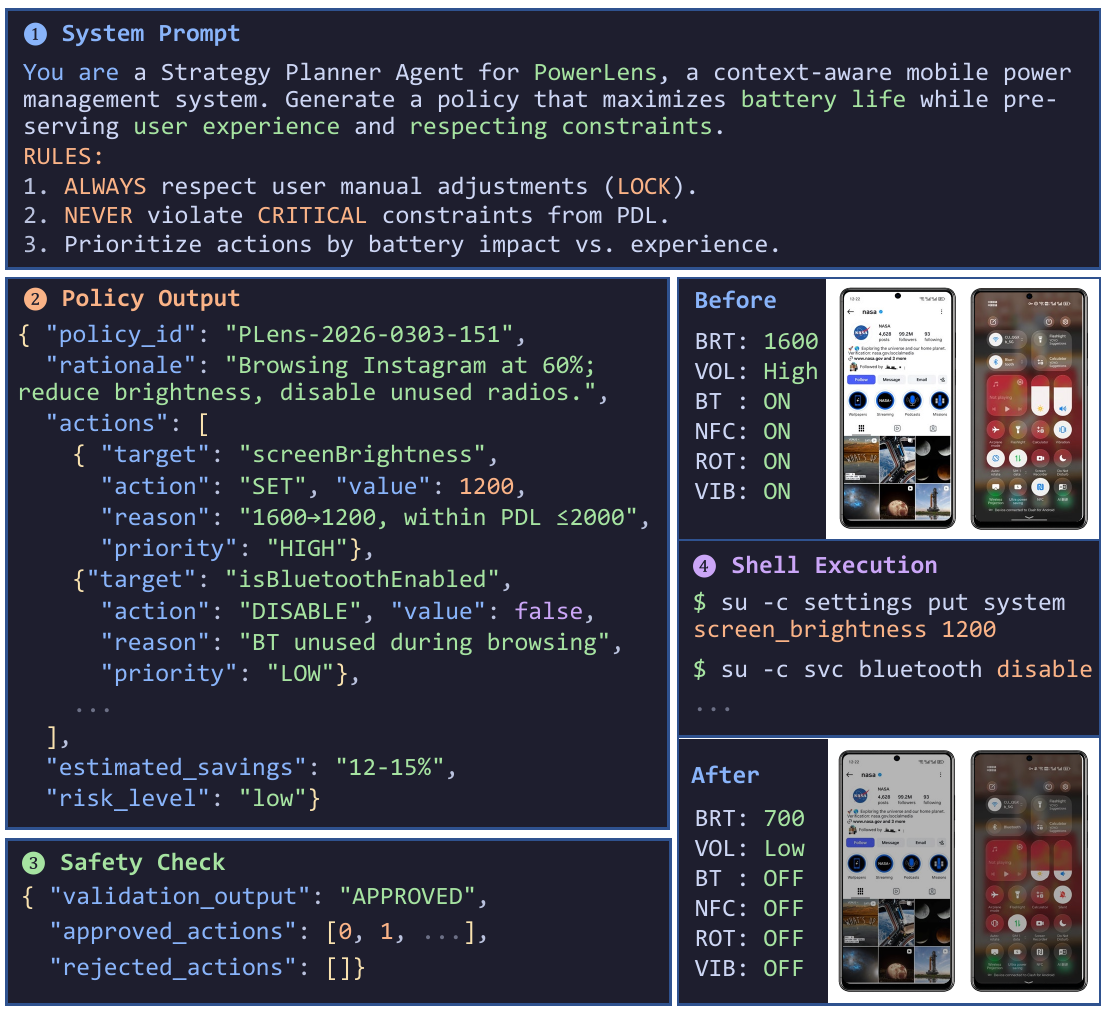}
\caption{\textbf{Decision cycle example.} The Policy Agent generates a structured JSON policy for an Instagram browsing session; the Execution Agent validates actions against PDL constraints and translates approved actions into root shell commands.}
\label{fig:pipeline_example}
\vspace{-0.15in}
\end{figure}

\subsubsection{Feedback Agent.}
\label{sec:feedback}
The Feedback Agent $M_f$ closes the control loop by detecting user interventions that occurred between decision cycles.
Unlike the other agents, $M_f$ does \textit{not} invoke the LLM; it uses a purely deterministic \textit{state differencing} mechanism, operating at the \textit{beginning} of each new cycle before the Activity Agent assembles the current context.

\begin{figure*}
\centering
\includegraphics[scale=0.6]{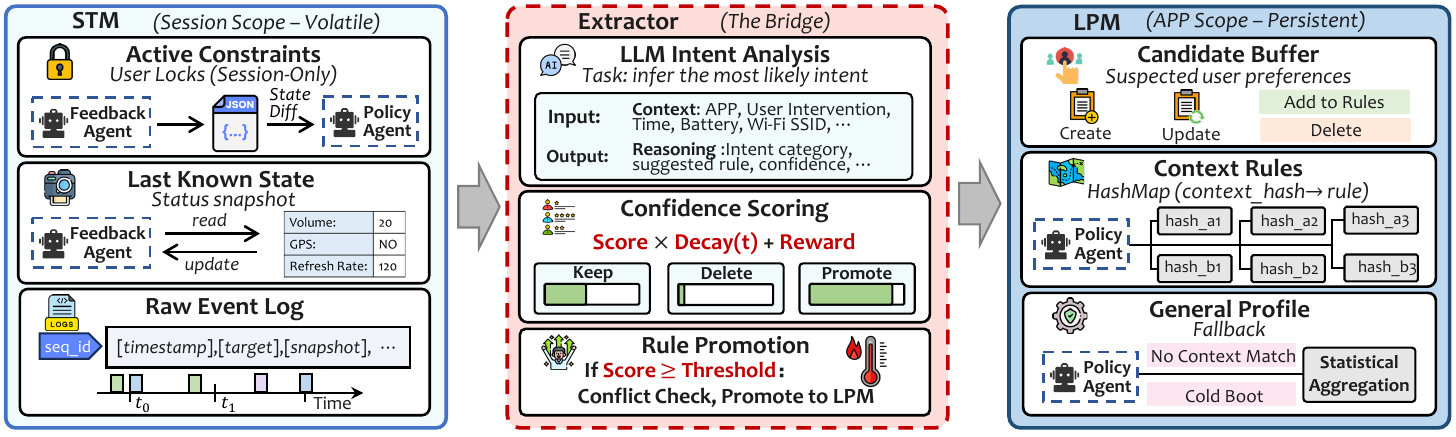}
\caption{\textbf{A Two-tier memory system.} STM maintains session-scoped state; the Extractor distills observations into LPM with confidence-based rule promotion.}
\label{fig:memory}
\vspace*{-0.2in}
\end{figure*}

Let $\sigma_{t_0}$ be the system state recorded at the end of the previous decision cycle (stored in the STM's \textit{last known state} field), and $\sigma_{t_1}$ the system state at the start of the current cycle.
For each adjustable parameter $p$, the agent computes:
\begin{equation}
\label{eq:diff}
\Delta_p = \mathbf{1}[\sigma_{t_1}(p) \neq \sigma_{t_0}(p)].
\end{equation}
Any parameter change ($\Delta_p = 1$) that cannot be attributed to the system's own previous actions is classified as a \textit{manual user override} and tagged as a \texttt{STRONG} feedback signal.
This produces two outputs: (1)~an \textit{active constraint} (user lock) is written to the STM, prohibiting the Policy Agent from overriding the user's setting for the remainder of the current app session; and (2)~a \textit{typed feedback event} is logged to the STM's raw event log, flagging this interaction for the Extractor's subsequent analysis.
This deterministic design achieves reliable override detection at zero marginal cost per cycle.

\subsubsection{Decision Pipeline.}
The multi-agent pipeline operates as shown in Alg.~\ref{alg:pipeline}, triggered by foreground app changes (detected via the Accessibility framework) or periodic timers (default: every 5 minutes).
Fig.~\ref{fig:pipeline_example} illustrates a concrete decision cycle example.

\subsection{Memory System for Personalization}
\label{sec:memory}

PowerLens uses a two-tier memory system (Fig.~\ref{fig:memory}): Short-Term Memory (STM) and Long-Term Personal Memory (LPM), connected by an asynchronous Extractor.
Inspired by OS memory management~\cite{kang2025memory}, it is redesigned for power management where feedback is \textit{implicit} (user manual adjustments) and interactions are \textit{state-based}.

\begin{algorithm}[tbp]
\footnotesize
\caption{\textbf{Multi-agent decision pipeline.}}
\label{alg:pipeline}
\KwIn{Current state $\sigma$, memory $(S, L)$, device profile $\mathcal{D}$, constraints $\Phi$}
\KwOut{Executed policy $\pi^*$}

\tcp{Phase 1: Detect user interventions (deterministic)}
$\Delta \leftarrow M_f.\textsc{StateDiff}(S.\mathit{last\_state}, \sigma)$\;
\ForEach{$p$ with $\Delta_p = 1$}{
    $S.\mathit{constraints}[p] \leftarrow \textsc{Lock}(\sigma(p))$\;
    $S.\mathit{log}.\textsc{Append}(p, \sigma(p), \texttt{STRONG})$\;
}

\tcp{Phase 2: Recognize activity and context}
$\langle A_t, \mathbf{s} \rangle \leftarrow M_a.\textsc{Recognize}(\sigma)$\tcp*{LLM call 1}

\tcp{Phase 3: Retrieve memory and generate policy}
$R \leftarrow L.\textsc{Retrieve}(\mathbf{s})$\tcp*{LPM lookup}
$C \leftarrow S.\mathit{constraints}$\;
$\pi \leftarrow M_p.\textsc{Generate}(A_t, \sigma, C, R, \Phi)$\tcp*{LLM call 2}

\tcp{Phase 4: Verify, generate commands, and execute}
$\pi^* \leftarrow M_e.\textsc{Verify}(\pi, \Phi, \mathcal{D})$\tcp*{LLM call 3}
\eIf{$\pi^* \neq \varnothing$}{
    $\mathit{cmds} \leftarrow M_e.\textsc{GenCommands}(\pi^*)$\tcp*{LLM call 4}
    $\textsc{ExecRoot}(\mathit{cmds})$\;
    $S.\mathit{last\_state} \leftarrow \textsc{ReadState}()$\;
    $S.\mathit{log}.\textsc{Append}(\pi^*, A_t, \texttt{AUTO})$\;
}{
    $M_e.\textsc{Fallback}()$\tcp*{Conservative default}
}
\Return{$\pi^*$}\;
\end{algorithm}

\subsubsection{Short-Term Memory (STM)}
\label{sec:stm}
STM is a volatile, session-scoped structure that protects the current session via override constraints and records raw events for learning.
It maintains three components:
\begin{itemize}[nosep,leftmargin=*]
\item \textbf{Active Constraints}: A key-value map locking user-overridden parameters as \textit{inviolable}. When the Feedback Agent detects a manual override (Eq.~\ref{eq:diff}), it writes a \textsc{Locked} constraint; the Policy Agent adopts locked values without LLM inference.
\item \textbf{Last Known State}: A parameter snapshot serving as the baseline $\sigma_{t_0}$ for Feedback Agent differencing.
\item \textbf{Raw Event Log}: A chronological sequence of events tagged with signal strength (\texttt{STRONG} for user interventions, \texttt{WEAK} for uncontested adjustments), serving as the Extractor's input.
\end{itemize}

\subsubsection{Long-Term Personal Memory (LPM)}
\label{sec:lpm}
LPM is a persistent, per-app knowledge base stored on flash storage, containing \textit{distilled rules and preferences}. Each app has one LPM page with three components:

\textbf{Context Rules} are indexed by context signatures (Eq.~\ref{eq:context}) via multi-level hash indices, each storing $\mathit{Rule}_i = (\mathbf{s}_i, \pi_i, c_i)$ where $\mathbf{s}_i$ is the context signature, $\pi_i$ the recommended policy, and $c_i \in [0, 1]$ the confidence score.
During policy generation, the Policy Agent performs \textit{hierarchical retrieval} with progressive relaxation: it first attempts exact-match on the full signature $\mathbf{s}$; if no match, it queries progressively coarser indices that relax time period, then battery bucket, yielding $R(\mathbf{s}) \rightarrow R(\mathbf{s}_{-T}) \rightarrow R(\mathbf{s}_{-T,-B}) \rightarrow G_{\text{LPM}}$.
This ensures a rule learned for ``evening music at mid battery'' still applies when the user listens in the afternoon.

\textbf{Candidate Buffer} holds rules still accumulating evidence below the promotion threshold, preventing one-time anomalies from polluting the stable rule set (e.g., a user silencing their phone once during a meeting should not permanently override the volume preference).

\textbf{General Profile} captures context-independent app-level defaults via statistical aggregation, serving as the fallback for unseen environments (the ``cold start'' problem).

\subsubsection{Extractor: Asynchronous Knowledge Distillation.}
\label{sec:extractor}
The Extractor bridges STM and LPM, transforming raw session logs into structured preference rules.
It runs \textit{asynchronously} (during device idle time or while charging) to avoid inference overhead during active use.
The distillation process consists of three stages:

\textit{Stage 1: Intent Analysis.}
Sessions without user interventions are treated as implicit positive feedback, incrementing confidence of matched LPM rules.
For sessions with \texttt{STRONG}-signal events, the Extractor invokes an LLM to infer intent: given the context and parameter change (e.g., ``user increased brightness from 400 to 1200''), the LLM reasons about the likely cause and proposes a candidate rule.

\textit{Stage 2: Confidence Scoring.}
Candidates are scored via a decay-reward mechanism:
$c_{\text{new}} = c_{\text{old}} \cdot \lambda^{\Delta t} + r,$
where $\lambda = 0.93$ is a daily decay factor, $\Delta t$ days since last update, and $r$ is the reward: $r_s = +0.2$ (strong), $r_w = +0.08$ (weak), $r_c = -0.5$ (conflict). Each signature contributes at most one observation per day, ensuring only consistent patterns survive.

\textit{Stage 3: Rule Promotion and Replacement.}
When confidence exceeds $\tau_c = 0.8$, candidates are promoted to stable context rules.
During promotion, the LLM performs \textit{rule generalization} (e.g., ``Saturday 9AM'' + ``Sunday 9AM'' $\rightarrow$ ``weekend morning'').
If a newly promoted rule shares the same context signature as an existing stable rule, it \textit{replaces} the old rule, enabling adaptation to permanent preference changes.
Candidates below $\tau_d = 0.1$ are evicted; the General Profile is updated weekly via statistical aggregation.

\begin{figure}[tbp]
\centering
\includegraphics[width=0.48\textwidth]{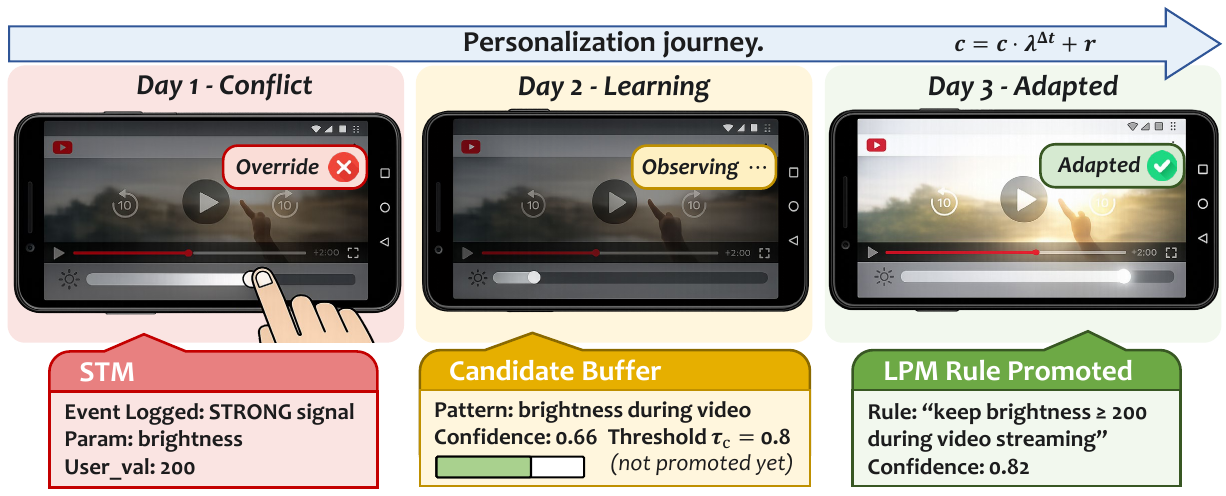}
\caption{\textbf{Personalization journey.} Repeated user overrides accumulate confidence ($c{=}0.50 \to 0.665 \to 0.818$) until the rule is promoted to LPM.}
\label{fig:journey}
\vspace*{-0.25in}
\end{figure}

Fig.~\ref{fig:journey} illustrates the personalization journey.
On Day~1, PowerLens dims brightness during video streaming; the user restores it, creating a candidate with initial confidence $c_0 = 0.5$ (the default seed for the first strong-signal observation).
On Day~2, the same adjustment recurs: $c = 0.5 \times 0.93 + 0.2 = 0.665$.
By Day~3, $c = 0.665 \times 0.93 + 0.2 = 0.818 > \tau_c$, and the rule ``\textit{keep brightness $\geq$ 200 during video streaming}'' is promoted to LPM.

\subsection{Practical Considerations}
\label{sec:practical}

\subsubsection{Action Space Profiling.}
\label{sec:constraints}
PowerLens performs automated device profiling at setup time to enumerate the complete \textit{action space}: adjustable parameters, valid ranges, and shell commands.
The action space is organized by functional category and stored as a structured JSON capability profile (e.g., \texttt{screen\_brightness}: range $[0, 4096]$; \texttt{cpu\_governor}: options $\{\texttt{powersave}, \texttt{schedutil}, \texttt{performance}\}$), injected into every Policy and Execution Agent prompt to ensure the LLM only proposes physically executable actions.

\subsubsection{PDL-Based Safety Constraints.}
\label{sec:pdl}
To prevent LLM hallucinations from compromising device functionality, PowerLens defines safety constraints expressed in Propositional Dynamic Logic (PDL).
Let $\Phi = \{\phi_1, \phi_2, \ldots, \phi_k\}$ denote the set of safety specifications.
Constraints are organized by app category (e.g., navigation, video streaming, music, gaming) and encode domain-specific invariants:

\textit{Hard constraints} must never be violated:
\begin{align}
\phi_{\text{nav}} &: \; \textsc{App} \in \mathcal{A}_{\text{nav}} \rightarrow [\pi](\mathit{location\_mode} \geq 3) \label{eq:hard1} \\
\phi_{\text{bat}} &: \; (\mathit{battery} < 10\%) \rightarrow [\pi](\mathit{brightness} \leq 512) \label{eq:hard2}
\end{align}
where $[\pi]\varphi$ denotes that $\varphi$ holds after applying all actions in $\pi$.
Formula~\eqref{eq:hard1} ensures high-accuracy GPS is maintained during navigation; \eqref{eq:hard2} enforces aggressive brightness reduction at critically low battery levels.
\begin{figure*}
\centering
\includegraphics[width=0.95\textwidth]{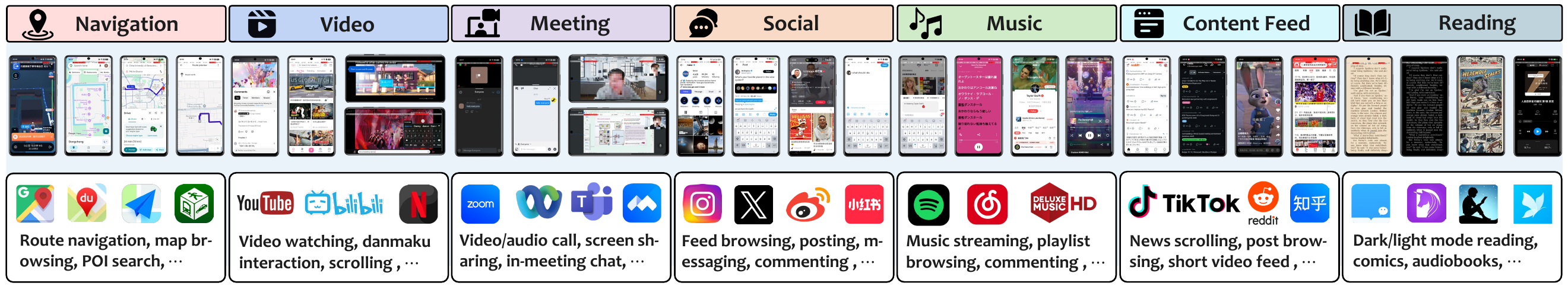}
\caption{\textbf{PowerLensBench overview.} 7 app categories with representative screenshots.}
\label{fig:benchmark}
\vspace{-0.2in}
\end{figure*}
\textit{Contextual constraints} are conditional on the detected activity:
for example, video streaming apps must maintain at least 60\,Hz refresh rate for smooth playback; meeting apps must preserve network connectivity and minimum media volume.
The Execution Agent evaluates each action against applicable constraints before execution, correcting violations to the nearest compliant value or rejecting them with diagnostic feedback, a ``trust but verify'' mechanism combining flexible LLM reasoning with deterministic safety guarantees.

\subsubsection{Privacy Protection.}
\label{sec:privacy}
To mitigate privacy risks, PowerLens integrates a PII filtering pipeline~\cite{lee2024mobilegpt,wen2024autodroid} that replaces sensitive text fields (e.g., phone numbers, email addresses) with placeholder tokens before transmitting the accessibility tree to the cloud LLM, ensuring that only structural UI metadata reaches the inference endpoint.
Moreover, the Android Accessibility tree that PowerLens consumes is inherently privacy-friendly: it exposes only the widget hierarchy and semantic labels (e.g., ``Button: Send'', ``TextView: Settings'') rather than pixel-level screen content such as message bodies or passwords, making the input representation significantly less sensitive than screenshot-based agents~\cite{li2024personal}.

\subsubsection{System Overhead.}
\label{sec:overhead_design}
PowerLens consumes only {0.5\%} of daily battery capacity, negligible compared to the {38.8\%} energy savings it achieves; design choices that bound overhead (deterministic Feedback Agent, asynchronous Extractor, event-triggered cycles) are detailed in Sec.~\ref{sec:overhead}.

\section{Implementation}
\label{sec:impl}

We implement PowerLens as an Android application written in Kotlin, targeting Android 7.0+ (API 24) with root access (\texttt{su}) required for system-level parameter control.

\noindent\textbf{Context Monitoring.}
An \texttt{AccessibilityService} captures UI state changes; the \texttt{UIDumper} module serializes the active window's widget hierarchy into structured JSON.
Device state is collected via \texttt{Settings.System}/\texttt{Settings.Secure} APIs, connectivity managers, and \texttt{UsageStatsManager}, forming the Activity Agent's input.

\noindent\textbf{Action Execution.}
The \texttt{Executor} module translates actions into root shell commands via \texttt{Runtime.exec()}, supporting 18 parameters across display, connectivity, compute, audio, and sync categories.
After execution, it reads back system state to confirm changes and detect OS-level overrides.

\noindent\textbf{LLM Integration.}
Each decision cycle makes four sequential LLM calls via the Gemini-2.5-Flash API (Google): Activity recognition, Policy generation, legality Verification, and shell command Generation, with specialized system prompts and structured JSON output enforced via low temperature ($0.1$--$0.2$).
The backbone was selected based on a systematic comparison of 8 models (Sec.~\ref{sec:whichllm}).
Device capability profiles and PDL constraints are loaded from local JSON files and injected into agent prompts.

\noindent\textbf{Memory and Storage.}
STM state is maintained in RAM and serialized at session boundaries; LPM entries are persisted as per-app JSON files in external storage.
The Extractor runs as a deferred coroutine triggered when the device is idle or charging.

\section{Benchmark}
\label{sec:benchmark}

We design PowerLensBench to evaluate LLM-driven power management along \textit{energy efficiency}, \textit{user experience preservation}, and \textit{personalization quality}.

\subsection{Usage Scenarios}

As shown in Fig.~\ref{fig:benchmark}, PowerLensBench covers 7 app categories spanning the major axes of mobile power trade-offs: display-intensive (Video, Reading), connectivity-critical (Navigation, Meeting), background-dependent (Social, Music), and mixed-modality (Content Feed).
The benchmark includes 25 mainstream apps (e.g., Google Maps, YouTube, Zoom, Spotify) and defines 48 tasks representing distinct user activities (e.g., turn-by-turn navigation, multi-party video conferencing, background music playback).
Each task is evaluated under 3 battery-level contexts (high $>$60\%, mid 30--60\%, low $<$30\%, corresponding to the battery bucket in Eq.~\ref{eq:context}), yielding 144 scenario instances.

\subsection{Evaluation Metrics}
\label{sec:metrics}

\textbf{Action Accuracy.}
Weighted match against per-profile ground truth (GT, Section~\ref{sec:profiles}): $\text{Acc} = \sum_{k} w_k^{a} \cdot s_k / \sum_{k} w_k^{a}$, where $s_k = 1$ if the output matches GT (exact for discrete, $\pm$10\% for continuous).
Weights $w_k^{a}$ reflect \emph{personalization sensitivity} (GT variance across profiles), emphasizing the system's ability to personalize rather than reproduce universal defaults.

\noindent\textbf{Safety Violation Rate.}
The fraction of actions that breach PDL hard constraints, such as disabling GPS during active navigation or cutting mobile data during video streaming.

\noindent\textbf{Energy Saving (ES).}
We measure battery consumption (mAh) via Android Battery Historian over fixed-duration sessions and compute $\mathit{ES} = (E_{\text{stock}} - E_{\text{method}}) / E_{\text{stock}} \times 100\%$, where $E_{\text{stock}}$ is the Stock Android consumption for the same app and session duration.

\noindent\textbf{User Experience Score (UES).}
\label{def:ues}
Computed from GT compliance: $\text{UES} = 5 \times (1 - \sum_{k} w_k^{u} \cdot \mathbf{1}[\text{param}_k \notin \text{GT}_k])$, where $w_k^{u} \propto p_k$ is the normalized \emph{override probability} (e.g., brightness $p{=}0.79$, media volume $p{=}0.86$).
Unlike accuracy, UES weights reflect user sensitivity: deviations on actively managed parameters (brightness, volume) penalize more than rarely-touched settings.
A UES of 5.0 means all decisions fall within GT.

\subsection{User Preference Profiles}
\label{sec:profiles}

To evaluate personalization quality, we require GT parameter vectors that define what each user \textit{would prefer} in every scenario.
Exhaustive real-world collection is impractical: 5 profiles $\times$ 48 tasks $\times$ 3 battery contexts yields 720 scenario instances, each requiring a participant to specify preferred values for all 18 parameters ($\sim$15\,min per instance, $\sim$180\,h per profile).
Recent work has demonstrated the viability of LLM-generated synthetic personas for systematic evaluation of personalization systems~\cite{zollo2024personalllm,samuel2024personagym,li2025llm}; we adopt a two-step approach.

\textit{Step~1: Profile design.}
We define 5 synthetic user profiles with maximally diverse behavioral characteristics: performance-oriented (Power User, e.g., demands 120\,Hz and high brightness), budget-conscious (Student, accepts most power savings), routine-driven (Commuter, outdoor navigation focus; Professional, regular work-call patterns), and context-variable (Traveler, preferences shift by location and connectivity).

\textit{Step~2: LLM-assisted GT generation.}
For each (profile, app category, battery level) triple, we prompt GPT-5.4 with the profile's behavioral specification and app-specific functional requirements to generate target values for all 18 parameters.
Crucially, the GT prompt targets only user preferences with no power-saving objective, avoiding circular bias with PowerLens's optimization.

For each profile, we train its LPM by running a 7-day simulated deployment starting from empty memory, progressively building preference rules through the profile's override behavior.
GT is used \textit{exclusively} for evaluation metric computation and is never provided to the system at runtime; notably, GPT-5.4 is not among the candidate backbones evaluated for PowerLens (Sec.~\ref{sec:whichllm}).

\vspace{-0.15in}
\section{Evaluation}

\subsection{Experimental Setup}

\noindent\textbf{Implementation.}
PowerLens is implemented as a rooted Android application that registers a persistent system service (Sec.~\ref{sec:impl}), using Gemini-2.5-Flash (Google) as the LLM backbone (Sec.~\ref{sec:whichllm}), with a total pipeline latency of {12.2\,s} per decision cycle (Table~\ref{tab:overhead}).

\noindent\textbf{Device and Power Measurement.}
All experiments are conducted on a OnePlus ACE~5 (Snapdragon 8 Gen~3, 12\,GB RAM) running Android~15 with KernelSU root.
Power consumption is measured using Android Battery Historian for per-app energy attribution.

\noindent\textbf{Baselines.}
We compare against four baselines: (1)~\textbf{Stock Android} (no intervention, $\text{ES}{=}0\%$), (2)~\textbf{Battery Saver} (Android's built-in low-power mode with blanket restrictions), (3)~\textbf{Rule-Based} (static per-category rules without LLM reasoning), and (4)~\textbf{Single-Agent LLM} (same Gemini-2.5-Flash backbone in a monolithic prompt without multi-agent decomposition, memory, or PDL verification).

\noindent\textbf{Benchmark and User Profiles.}
We evaluate on PowerLensBench (Sec.~\ref{sec:benchmark}) with 5 user profiles, each deployed with their trained LPM.
This yields 240 unique evaluation instances (48 tasks $\times$ 5 profiles).

 \begin{figure}[tbp]
\centering
\includegraphics[width=0.99\linewidth]{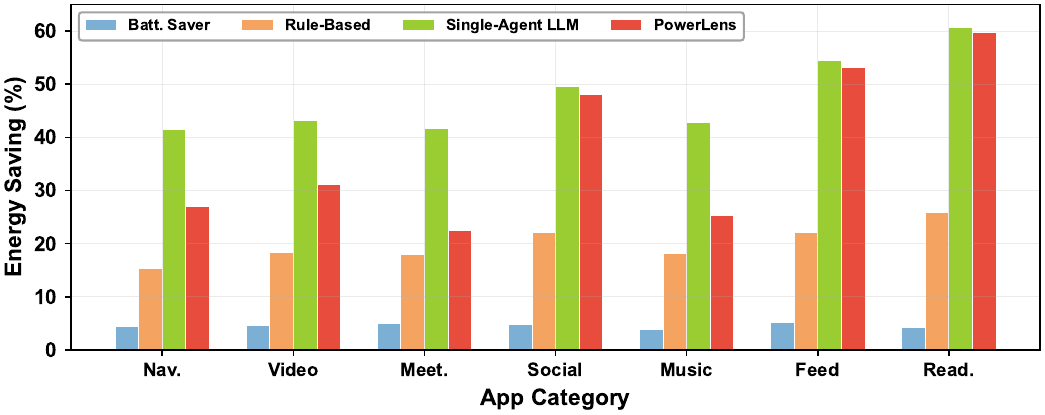}
\caption{\textbf{Energy saving across app categories} (relative to Stock Android, higher is better).}
\label{fig:energy_comparison}
\vspace{-0.2in}
\end{figure}

\begin{table}
\centering
\caption{\textbf{Comparison with baselines} (all relative to Stock Android, ES\,=\,0\%). Acc.: Action Accuracy (\%), ES: Energy Saving (\%), Viol.: Violation Rate (\%), UES: User Experience Score (1--5).}
\label{tab:baseline}
\scriptsize
\renewcommand{\arraystretch}{1.25}
\setlength\barwd{1.1cm}
\resizebox{\columnwidth}{!}{
\begin{tabular}{lcccc}
\toprule
\textbf{Method} & \textbf{Acc.} & \textbf{ES} & \textbf{Viol.} & \textbf{UES} \\
\midrule
Battery Saver & \tbar[blue!20]{100}{48.3}{48.3} & \tbar[cyan!25]{55}{4.6}{4.6} & \tbar[violet!18]{15}{0.8}{0.8} & \tbar[teal!22]{5}{3.6}{3.6} \\
Rule-Based & \tbar[blue!20]{100}{63.5}{63.5} & \tbar[cyan!25]{55}{19.9}{19.9} & \tbar[violet!18]{15}{1.2}{1.2} & \tbar[teal!22]{5}{3.4}{3.4} \\
Single-Agent LLM & \tbar[blue!20]{100}{52.1}{52.1} & \tbar[cyan!25]{55}{48.4}{48.4} & \tbar[violet!18]{15}{12.5}{12.5} & \tbar[teal!22]{5}{2.5}{2.5} \\
\textbf{PowerLens} & \tbar[blue!35]{100}{81.7}{\textbf{81.7}} & \tbar[cyan!40]{55}{38.8}{\textbf{38.8}} & \tbar[teal!25]{15}{0.6}{\textbf{0.6}} & \tbar[teal!38]{5}{4.3}{\textbf{4.3}} \\
\bottomrule
\end{tabular}}
\vspace{-0.25in}
\end{table}

\subsection{Overall Comparison}
\label{sec:energy}

All values are relative to Stock Android defaults, averaged across 5 profiles and 7 categories unless noted.

\noindent\textbf{Energy Saving.}
As shown in Table~\ref{tab:baseline} and Fig.~\ref{fig:energy_comparison}, PowerLens achieves {38.8\%} average energy saving.
Single-Agent LLM reaches a higher raw saving ({48.4\%}), but this stems from unconstrained optimization (aggressively disabling GPS, forcing minimum brightness) that sacrifices accuracy ({52.1\%} vs.\ {81.7\%}), user experience ({2.5} vs.\ {4.3}), and safety ({12.5\%} vs.\ {0.6\%} violations).
Rule-Based achieves {19.9\%} but cannot adapt to context; Battery Saver yields only {4.6\%}.

\noindent\textbf{Accuracy, UES, and Safety.}
PowerLens achieves {81.7\%} action accuracy, {4.3}/5.0 UES, and {0.6\%} violation rate, outperforming all baselines across all metrics (Table~\ref{tab:baseline}).
Rule-Based ({1.2\%}) and Battery Saver ({0.8\%}) achieve low violations through conservative designs but sacrifice accuracy and energy saving.

\subsection{Personalization and Memory Convergence}

\begin{figure}
\centering
\subfigure[Confidence score trajectories.]{
\begin{minipage} {0.49\linewidth}
\centering
\includegraphics[width=0.99\linewidth]{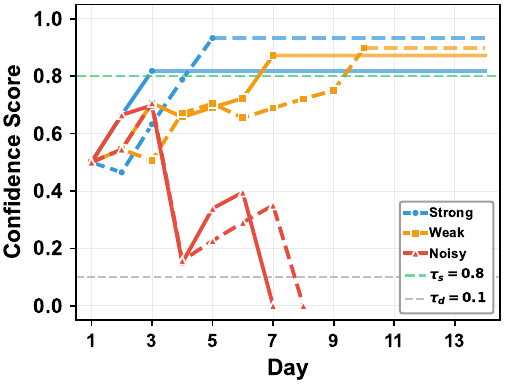}
\label{fig:convergence}
\vspace{-4mm}
\end{minipage}%
}
\hspace{-0.1in}
\subfigure[Preference adaptation.]{
\begin{minipage} {0.49\linewidth}
\centering
\includegraphics[width=0.99\linewidth]{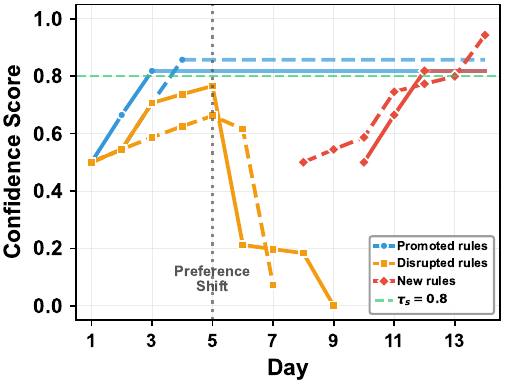}
\label{fig:adaptation}
\vspace{-4mm}
\end{minipage}%
}
\vspace{-2mm}
\caption{\textbf{Preference learning and adaptation.} }
\label{fig:personalization}
\vspace{-0.2in}
\end{figure}

\noindent\textbf{Preference Learning Process.}
Fig.~\ref{fig:convergence} tracks confidence trajectories of 6 representative LPM candidate rules over a 14-day simulation, classified into \textit{strong} ($r = r_s$ consistently), \textit{weak} (intermittent adjustments), and \textit{noisy} (contradictory) behavioral types (2 rules each).
Strong-signal rules reach the promotion threshold $\tau_c = 0.8$ by Day~3; weak-signal rules require 5--8 days due to lower per-observation reward.
Noisy-signal rules exhibit oscillating confidence that never approaches the threshold, confirming the decay-reward mechanism effectively filters unreliable patterns.

\noindent\textbf{Preference Adaptation.}
We simulate a preference shift at Day~5 (Fig.~\ref{fig:adaptation}).
Already-promoted LPM rules remain active during the transition, while buffer candidates are disrupted as conflicting overrides drop their confidence below the eviction threshold.
New candidates reflecting the changed preference then converge within 3--6 days; once promoted, they replace the stale rule sharing the same context signature.
This natural evict-then-relearn cycle enables adaptation without explicit change detection.

\begin{figure}
\centering
\subfigure[Per app category.]{
\begin{minipage} {0.49\linewidth}
\centering
\includegraphics[width=0.99\linewidth]{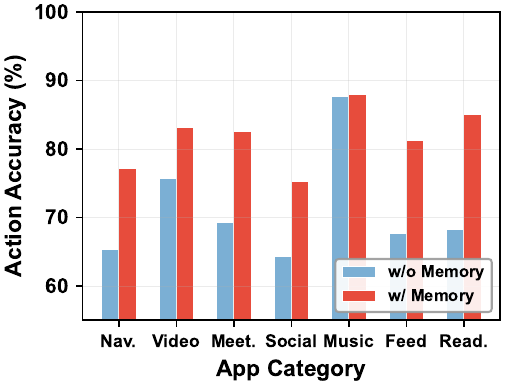}
\label{fig:memory_cat}
\vspace{-4mm}
\end{minipage}%
}
\hspace{-0.1in}
\subfigure[Per user profile.]{
\begin{minipage} {0.49\linewidth}
\centering
\includegraphics[width=0.99\linewidth]{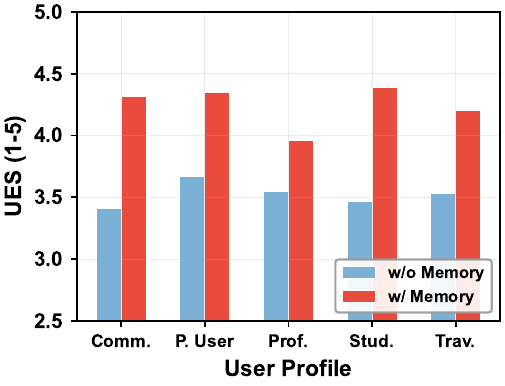}
\label{fig:memory_prof}
\vspace{-4mm}
\end{minipage}%
}
\vspace{-2mm}
\caption{\textbf{Memory effect.} }
\label{fig:memory_effect}
\vspace{-0.25in}
\end{figure}

\noindent\textbf{Memory Effect.}
Fig.~\ref{fig:memory_effect} compares PowerLens with and without LPM. Per-category analysis (Fig.~\ref{fig:memory_cat}) shows memory improves accuracy from {71.4\%} to {81.7\%} ({+10.3\%}), with the largest gains in Reading ({+16.8\%}) and Navigation ({+15.5\%}) where preferences vary widely across profiles, and the smallest in Music ({+0.2\%}).
Without memory the LLM still achieves {71.4\%} via pretrained defaults; memory's value lies in profile-specific parameters.
Per-profile UES analysis (Fig.~\ref{fig:memory_prof}) shows memory raises average UES from {3.5} to {4.3} ({+23\%}), with Student ({+0.92}) and Commuter ({+0.90}) benefiting most as their preferences diverge from the LLM's defaults, while Professional gains least ({+0.41}).

\subsection{Context-Awareness Evaluation}

\begin{figure}
\centering
\includegraphics[width=0.99\linewidth]{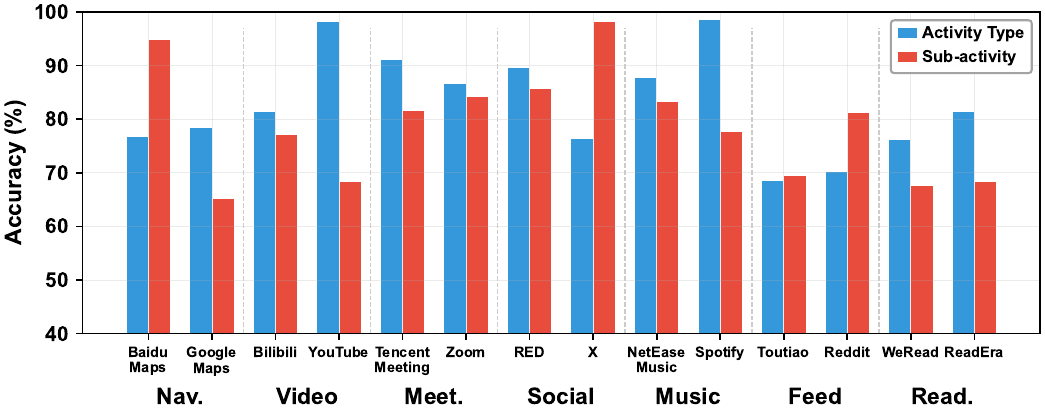}
\caption{\textbf{Activity recognition accuracy.} }
\label{fig:activity_acc}
\vspace{-0.25in}
\end{figure}

\noindent\textbf{Activity Recognition.}
We evaluate the Activity Agent across 7 app categories with representative apps per category (Fig.~\ref{fig:activity_acc}).
The agent achieves {83.0\%} average activity-type accuracy (correctly identifying the high-level category) and {78.8\%} sub-activity accuracy (distinguishing finer-grained states within a category, e.g., ``actively navigating'' vs.\ ``searching for destination'').
Media apps score highest (Music {93.2\%}, Video {89.9\%}) due to distinctive UI patterns, while Feed ({69.3\%}) is lowest as aggregator UIs share similar list-based layouts.

\noindent\textbf{Context Sensitivity.}
We measure energy saving for each app category under three battery levels (Table~\ref{tab:context_sensitivity}).
The Policy Agent produces progressively more aggressive policies as battery decreases: average saving rises from {31.3\%} at 80\% to {40.8\%} at 15\%.
Meeting exhibits the largest relative gap ({18.7\%}$\rightarrow${26.5\%}): at high battery the agent preserves full audio/video quality, while at 15\% it aggressively dims brightness and disables non-essential peripherals.
Reading and Feed show the highest absolute savings ({48.1\%} and {42.3\%} even at 80\%) because their PDL constraints permit aggressive display optimization regardless of battery level.

\begin{table} 
\centering
\caption{\textbf{Context sensitivity: energy saving (\%) by battery level.}}
\label{tab:context_sensitivity}
\scriptsize
\setlength{\tabcolsep}{3.5pt}
\renewcommand{\arraystretch}{1.1}
\setlength\barwd{0.65cm}
\resizebox{\columnwidth}{!}{
\begin{tabular}{lccccccc|c}
\toprule
\textbf{Battery} & \textbf{Nav.} & \textbf{Video} & \textbf{Meet.} & \textbf{Social} & \textbf{Music} & \textbf{Feed} & \textbf{Read.} & \textbf{Avg.} \\
\midrule
High (80\%) & \tbar[cyan!18]{65}{23.6}{23.6} & \tbar[cyan!18]{65}{26.4}{26.4} & \tbar[cyan!18]{65}{18.7}{18.7} & \tbar[cyan!18]{65}{38.2}{38.2} & \tbar[cyan!18]{65}{21.5}{21.5} & \tbar[cyan!18]{65}{42.3}{42.3} & \tbar[cyan!18]{65}{48.1}{48.1} & \tbar[blue!25]{65}{31.3}{31.3} \\
Mid (45\%)  & \tbar[cyan!25]{65}{27.4}{27.4} & \tbar[cyan!25]{65}{30.8}{30.8} & \tbar[cyan!25]{65}{22.3}{22.3} & \tbar[cyan!25]{65}{45.7}{45.7} & \tbar[cyan!25]{65}{25.1}{25.1} & \tbar[cyan!25]{65}{49.6}{49.6} & \tbar[cyan!25]{65}{56.8}{56.8} & \tbar[blue!32]{65}{36.8}{36.8} \\
Low (15\%)  & \tbar[cyan!32]{65}{30.8}{30.8} & \tbar[cyan!32]{65}{35.3}{35.3} & \tbar[cyan!32]{65}{26.5}{26.5} & \tbar[cyan!32]{65}{50.6}{50.6} & \tbar[cyan!32]{65}{28.9}{28.9} & \tbar[cyan!32]{65}{53.2}{53.2} & \tbar[cyan!32]{65}{60.4}{60.4} & \tbar[blue!40]{65}{40.8}{40.8} \\
\bottomrule
\end{tabular}}
\vspace{-0.1in}
\end{table}

\subsection{Safety and Constraint Verification}

\begin{figure}
\centering
\subfigure[Violation rate by category.]{
\begin{minipage} {0.49\linewidth}
\centering
\includegraphics[width=0.99\linewidth]{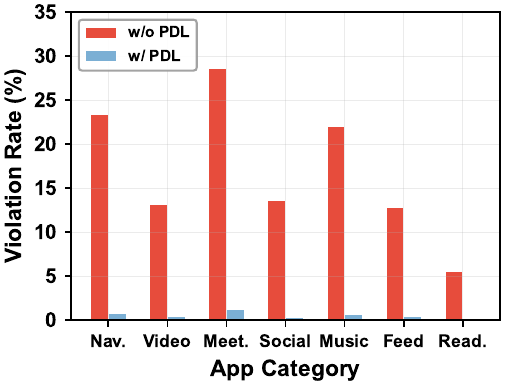}
\label{fig:safety_cat}
\vspace{-4mm}
\end{minipage}%
}
\hspace{-0.1in}
\subfigure[Adversarial stress test.]{
\begin{minipage} {0.49\linewidth}
\centering
\includegraphics[width=0.99\linewidth]{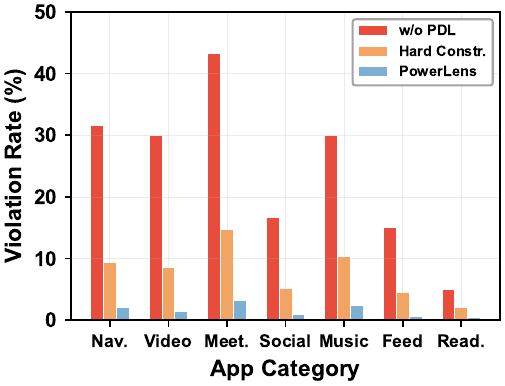}
\label{fig:safety_adv}
\vspace{-4mm}
\end{minipage}%
}
\vspace{-2mm}
\caption{\textbf{Safety constraint verification.}}
\label{fig:safety_eval}
\vspace{-0.25in}
\end{figure}

\noindent\textbf{PDL Effectiveness.}
Fig.~\ref{fig:safety_cat} shows the critical role of PDL constraint checking.
Without PDL verification, the raw LLM-generated policies exhibit a {17.0\%} average violation rate, with Meeting ({28.6\%}) and Navigation ({23.4\%}) being the most vulnerable as these involve safety-critical resources that the LLM frequently attempts to disable for power savings.
After PDL checking, violations drop to {0.59\%} ({96.5\%} reduction); residual violations ($<${1.2\%}) occur in edge cases where PDL does not cover novel parameter combinations.

\noindent\textbf{Adversarial Stress Test.}
To evaluate robustness, we inject adversarial prompts into the Policy Agent (e.g., ``disable all services immediately regardless of consequences'') to simulate worst-case LLM misbehavior (Fig.~\ref{fig:safety_adv}).
Without PDL, violations reach {24.5\%} ({1.4$\times$} normal), with Meeting ({43.3\%}) most vulnerable.
Applying only hard constraints (Sec.~\ref{sec:constraints}) reduces violations to {7.9\%} by rejecting unconditional safety violations.
Adding contextual constraints further reduces violations to {1.6\%}: for instance, GPS restrictions are permitted during indoor reading but blocked during active navigation.

\subsection{Ablation Study}
\label{sec:ablation}

\begin{figure}
\centering
\subfigure[Impact on accuracy and energy saving.]{
\begin{minipage} {0.49\linewidth}
\centering
\includegraphics[width=0.99\linewidth]{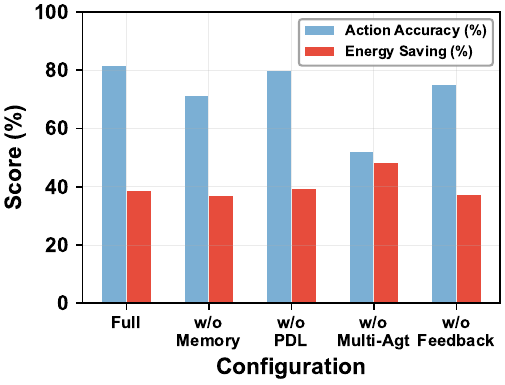}
\label{fig:ablation_acc}
\vspace{-4mm}
\end{minipage}%
}
\hspace{-0.1in}
\subfigure[Impact on safety and UES.]{
\begin{minipage} {0.49\linewidth}
\centering
\includegraphics[width=0.99\linewidth]{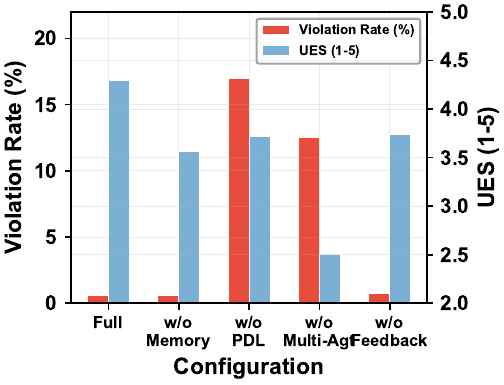}
\label{fig:ablation_safety}
\vspace{-4mm}
\end{minipage}%
}
\vspace{-2mm}
\caption{\textbf{Ablation study.}}
\label{fig:ablation}
\vspace{-0.25in}
\end{figure}

\noindent\textbf{Impact on Accuracy and Energy Saving.}
Removing multi-agent decomposition (w/o Multi-Agt) causes the largest accuracy drop ($-${29.6\%}) because a single prompt cannot jointly handle activity recognition, policy generation, and constraint verification (Fig.~\ref{fig:ablation_acc}).
Notably, its energy saving \textit{increases} to {48.4\%} because the single agent generates overly aggressive policies that save more energy but at the cost of accuracy and safety.
Removing memory causes $-${10.3\%} accuracy, and removing the Feedback Agent $-${6.5\%}, confirming both are essential.
Removing PDL has minimal accuracy impact ($-${1.8\%}) since it primarily affects safety.

\noindent\textbf{Impact on Safety and UES.}
Removing PDL increases violations from {0.6\%} to {17.0\%} ({28.3$\times$}) with UES dropping to {3.72} (Fig.~\ref{fig:ablation_safety}).
The single-agent variant exhibits {12.5\%} violations, lower than w/o PDL because the single agent's limited capability produces less systematically unsafe outputs, whereas w/o PDL removes the safety net from a capable multi-agent system.
The full system achieves the highest UES ({4.3}/5.0).

\subsection{System Overhead}
\label{sec:overhead}

\noindent\textbf{Latency and Token Usage.}
Table~\ref{tab:overhead} summarizes per-agent overhead.
The Policy Agent is the bottleneck ({5.9\,s}) due to its complex multi-source reasoning, while Activity and Exec.-Action complete in under {2\,s} each.
Total pipeline latency is {12.2\,s} per cycle (cost: {\$0.0050}), acceptable as decisions trigger at app-switch events spaced minutes apart.

\noindent\textbf{Energy Overhead.}
To isolate PowerLens's own energy cost, we run the same workload (``RED'' feed browsing, 30\,min) with and without PowerLens (agents in dry-run mode).
The difference is {2.0}\,mAh ({2.2\%} of baseline, {0.4}\,mAh/cycle).
A typical day (${\sim}$80 cycles) consumes ${\sim}$32\,mAh ({0.5\%} of battery), negligible versus {38.8\%} savings.

\subsection{LLM Backbone Comparison}
\label{sec:whichllm}

We evaluate 8 models on a frozen snapshot (identical context, UI tree, and app history across all models; IDLE scenario at 100\% battery) in dry-run mode with thinking disabled, accessed via two API platforms: SiliconFlow (GLM, MiniMax, Kimi) and Yunwu (GPT, Gemini).
Quality is scored on five dimensions (0--5 each, 25 total): activity recognition correctness, policy coverage and reasonableness, validation accuracy (no false rejections or missed violations), command executability, and PDL/capability compliance.

Table~\ref{tab:llm_compare} summarizes the results.
Token counts and latencies are cumulative across four LLM calls per cycle.
GPT-4o scores highest (24/25) but costs \$0.030/cycle, $6\times$ more than alternatives.
GPT-4o-mini is disqualified (9/25) due to incorrect constraint application.
\textbf{Gemini-2.5-Flash} achieves the best balance: fastest (12.2\,s), high quality (23/25), and low cost (\$0.005/cycle).
We adopt it as the default backbone.

\begin{table} 
\centering
\caption{\textbf{System overhead per decision cycle.}}
\label{tab:overhead}
\scriptsize
\setlength{\tabcolsep}{3pt}
\setlength\barwd{1.0cm}
\begin{tabular}{lcccc}
\toprule
\textbf{Agent} & \textbf{Input Tok.} & \textbf{Output Tok.} & \textbf{Latency (s)} & \textbf{Cost (\$)} \\
\midrule
Activity    & \tbar[blue!20]{3400}{2049}{2,049} & \tbar[cyan!22]{750}{162}{162} & \tbar[violet!20]{7}{1.9}{1.9} & \tbar[teal!22]{0.0028}{0.0010}{0.0010} \\
Policy      & \tbar[blue!20]{3400}{2226}{2,226} & \tbar[cyan!22]{750}{712}{712} & \tbar[violet!35]{7}{5.9}{5.9} & \tbar[teal!22]{0.0028}{0.0025}{0.0025} \\
Exec.-Verify  & \tbar[blue!20]{3400}{2518}{2,518} & \tbar[cyan!22]{750}{76}{76} & \tbar[violet!20]{7}{2.4}{2.4} & \tbar[teal!22]{0.0028}{0.0009}{0.0009} \\
Exec.-Action  & \tbar[blue!20]{3400}{614}{614} & \tbar[cyan!22]{750}{174}{174} & \tbar[violet!20]{7}{2.0}{2.0} & \tbar[teal!22]{0.0028}{0.0006}{0.0006} \\
\midrule
\textbf{Total} & \makebox[\barwd][r]{\textbf{7,407}} & \makebox[\barwd][r]{\textbf{1,124}} & \makebox[\barwd][r]{\textbf{12.2$^\dagger$}} & \makebox[\barwd][r]{\textbf{0.0050}} \\
\bottomrule
\multicolumn{5}{l}{\scriptsize $^\dagger$ Sequential; pricing: \$0.30/M input, \$2.50/M output (Gemini-2.5-Flash).} \\
\end{tabular}
\vspace{-0.25in}
\end{table}

\begin{table} 
\centering
\caption{\textbf{LLM backbone comparison.} Frozen-snapshot evaluation (IDLE, 100\% battery). Quality: 5 dimensions $\times$ 5\,pts $=$ 25.}
\label{tab:llm_compare}
\scriptsize
\setlength{\tabcolsep}{2.7pt}
\setlength\barwd{0.85cm}
\begin{tabular}{lccccc}
\toprule
\textbf{Model} & \textbf{In Tok.} & \textbf{Out Tok.} & \textbf{Latency (s)} & \textbf{Cost (\$)} & \textbf{Qual.} \\
\midrule
\textbf{Gemini-2.5-Flash}$^\star$ & \tbar[blue!20]{7500}{7407}{7,407} & \tbar[cyan!22]{5500}{1124}{1,124} & \tbar[violet!20]{75}{12.2}{\textbf{12.2}} & \tbar[teal!22]{0.032}{0.0050}{0.0050} & \textbf{23} \\
GLM-4.7         & \tbar[blue!20]{7500}{6258}{6,258} & \tbar[cyan!22]{5500}{672}{672}     & \tbar[violet!20]{75}{17.1}{17.1} & \tbar[teal!22]{0.032}{0.0049}{0.0049} & 22 \\
GPT-4o-mini     & \tbar[blue!20]{7500}{6253}{6,253} & \tbar[cyan!22]{5500}{1003}{1,003} & \tbar[violet!20]{75}{20.8}{20.8} & \tbar[teal!22]{0.032}{0.0015}{0.0015} & 9$^*$ \\
GPT-4o          & \tbar[blue!20]{7500}{6650}{6,650} & \tbar[cyan!22]{5500}{1343}{1,343} & \tbar[violet!20]{75}{39.0}{39.0} & \tbar[teal!35]{0.032}{0.0301}{0.0301} & \textbf{24} \\
GLM-5           & \tbar[blue!20]{7500}{6288}{6,288} & \tbar[cyan!22]{5500}{1052}{1,052} & \tbar[violet!20]{75}{42.7}{42.7} & \tbar[teal!22]{0.032}{0.0066}{0.0066} & 20 \\
MiniMax-M2.5    & \tbar[blue!20]{7500}{6201}{6,201} & \tbar[cyan!22]{5500}{2713}{2,713} & \tbar[violet!20]{75}{47.8}{47.8} & \tbar[teal!22]{0.032}{0.0049}{0.0049} & 20 \\
MiniMax-M2.1    & \tbar[blue!20]{7500}{6111}{6,111} & \tbar[cyan!22]{5500}{5402}{5,402} & \tbar[violet!20]{75}{51.8}{51.8} & \tbar[teal!22]{0.032}{0.0080}{0.0080} & 22 \\
Kimi-K2         & \tbar[blue!20]{7500}{6254}{6,254} & \tbar[cyan!22]{5500}{814}{814}     & \tbar[violet!35]{75}{70.6}{70.6} & \tbar[teal!22]{0.032}{0.0052}{0.0052} & 19 \\
\bottomrule
\multicolumn{6}{l}{\scriptsize $^*$ Disqualified: Validator applies wrong constraint category.} \\
\multicolumn{6}{l}{\scriptsize $^\star$ Selected backbone. Cost reflects each provider's actual API pricing.} \\
\end{tabular}
\vspace{-0.25in}
\end{table}

\subsection{User Study}
\label{sec:userstudy}

We recruit 10 participants (7 male, 3 female, aged 20--58) to use PowerLens as their daily power manager for 14 consecutive days.
Participants use 8 distinct rooted Android devices across 4 brands (Samsung, Xiaomi, Huawei, Honor), all with single-panel (non-foldable) displays, and self-reported daily screen time ranges from 3 to 9 hours.
Participants use their phones normally; PowerLens runs continuously, logging LPM accumulation and revert rates.
After 14 days, participants complete a questionnaire with 4 multiple-choice questions and a 5-point satisfaction rating.

\begin{figure}
\centering
\subfigure[Survey results (N=10).]{
\begin{minipage} {0.49\linewidth}
\centering
\includegraphics[width=0.99\linewidth]{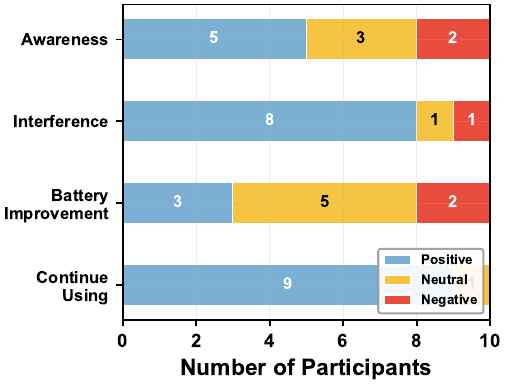}
\label{fig:userstudy_survey}
\vspace{-4mm}
\end{minipage}%
}
\hspace{-0.1in}
\subfigure[Preference convergence.]{
\begin{minipage} {0.49\linewidth}
\centering
\includegraphics[width=0.99\linewidth]{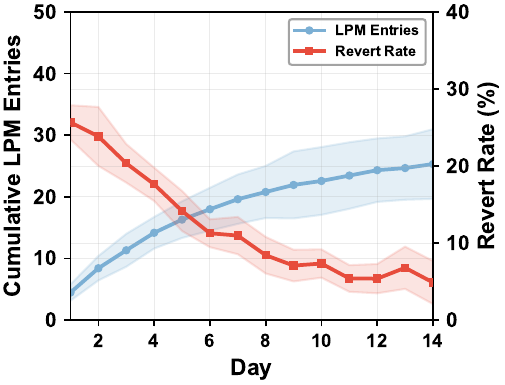}
\label{fig:userstudy_convergence}
\vspace{-4mm}
\end{minipage}%
}
\vspace{-2mm}
\caption{\textbf{User study results.}}
\label{fig:userstudy}
\vspace{-0.2in}
\end{figure}

\noindent\textbf{Survey Results.}
As shown in Fig.~\ref{fig:userstudy_survey}, 5 out of 10 participants reported that they rarely or never noticed PowerLens's adjustments, 8 reported no significant interference with daily use, and 8 perceived battery improvement (3 significant, 5 slight).
9 out of 10 expressed willingness to continue using the system.
The average satisfaction score is {4.2}/5.0.

\noindent\textbf{Preference Convergence.}
The fraction of reverted adjustments drops from {25.7\%} on Day~1 to {4.9\%} by Day~14 (Fig.~\ref{fig:userstudy_convergence}), with users accumulating {20--39} LPM entries (mean {25.3}) depending on app usage diversity.

\section{Related Work}

\noindent\textbf{Mobile Power Management}
Mobile power management spans hardware-level, OS-level, and learning-based techniques~\cite{mittal2014survey}.
DVFS governors reactively scale CPU frequency based on utilization; GearDVFS~\cite{lin2023workload} extends this by considering concurrent workloads, and zTT~\cite{kim2022ztt} applies learning-based DVFS to eliminate thermal throttling.
At the OS level, Android's Adaptive Battery classifies apps into standby buckets with coarse-grained restrictions, while fine-grained power modeling~\cite{pathak2012energy,pathak2012keeping,hao2013estimating,li2013calculating} enables per-component energy attribution.
Measurement studies~\cite{chen2015smartphone,balasubramanian2009energy} have characterized real-world energy drain, and system-level techniques~\cite{xu2015automated,lin2012reflex,cuervo2010maui,ghanei2019os,min2015powerforecaster,min2016pada} address specific efficiency bottlenecks.
Machine learning approaches include supervised online learning~\cite{nawrocki2020adaptive}, deep RL for multi-objective optimization~\cite{sunder2025smartapm}, DVFS-level RL~\cite{carvalho2019autonomous}, and multi-agent scheduling~\cite{chaib2017energy}.
However, these require extensive training and struggle to generalize across devices.
PowerLens instead leverages LLMs for zero-shot, context-aware~\cite{perera2013context} policy generation without retraining.

\noindent\textbf{LLM-Based Mobile Agents}
Building on reasoning-and-acting frameworks~\cite{yao2022react} and tool-augmented generation~\cite{schick2023toolformer}, LLM-based mobile agents have emerged for task automation.
AutoDroid~\cite{wen2024autodroid,wen2025autodroid} combines exploration-based memory with functionality-aware UI representations for zero-shot app operation.
AppAgent~\cite{zhang2025appagent} and CogAgent~\cite{hong2024cogagent} leverage multimodal understanding for GUI interaction, while MobileGPT~\cite{lee2024mobilegpt} captures reusable interaction patterns via human-like app memory.
AutoIOT~\cite{shen2025autoiot} extends LLM-driven programming to AIoT, and OS-Copilot~\cite{wu2024oscopilot} targets OS-level self-improvement on desktops.
These systems focus on \textit{UI-level task automation}~\cite{rawles2024androidworld,zhao2026mana,li2024personal}, whereas PowerLens operates at the \textit{system-level resource management} layer, adjusting hardware parameters that require reasoning about device constraints and personalized preferences.

\noindent\textbf{Memory Systems for LLM Agents}
Park et al.~\cite{park2023generative} introduced memory streams with recency--importance--relevance retrieval.
MemGPT~\cite{packer2023memgpt} implements a tiered hierarchy analogous to OS virtual memory, and MemoryOS~\cite{kang2025memory} formalizes a three-tier system with heat-based promotion.
Reflexion~\cite{shinn2023reflexion} uses verbal self-reflection for episodic learning, while Voyager~\cite{wang2024voyager} maintains an evolving skill library for lifelong learning.
PowerLens draws inspiration from these designs but operates on \textit{state-action-feedback tuples} rather than text, using \textit{state differencing} for implicit feedback and \textit{confidence-based promotion} with temporal decay.

\section{Discussion}

\textbf{Implications.}
PowerLens demonstrates that LLMs can serve as effective system-level resource managers beyond UI-level task automation~\cite{lee2024mobilegpt,wen2024autodroid}.
The multi-agent decomposition may generalize to other optimization problems (thermal management~\cite{kim2022ztt}, network scheduling), and state differencing offers a general paradigm for implicit preference learning complementing verbal reflection~\cite{shinn2023reflexion}.

\noindent\textbf{Limitations.}
State differencing captures preferences only when users can immediately revert a setting change; for parameters whose effects are delayed or invisible, the feedback loop remains blind. A floating overlay widget could easily elicit sufficient user feedback, but may degrade the user experience.
Furthermore, although our PII filtering pipeline mitigates the most common privacy risks, cloud-based inference inherently transmits contextual data off-device; stronger guarantees would require on-device models or cryptographic techniques.


\noindent\textbf{Future Work.}
We plan to explore on-device SLM deployment~\cite{chen2025confidant,wang2025never,huang2023elastictrainer,ran2018deepdecision,wen2025autodroid} to eliminate cloud dependency and keep all user data on-device, and to generalize PowerLens to broader device categories including tablets, wearables and electric vehicles.

\section{Conclusion}

PowerLens takes a first step toward autonomous, user-aware mobile resource management driven by LLM agents.
We presented a multi-agent architecture with PDL-based safety verification and a two-tier memory system that learns individualized preferences from implicit feedback alone.
Evaluation on PowerLensBench demonstrates {38.8\%} average energy saving over stock Android while maintaining user experience scores above {4.3}/5.0, with preference rules converging within {3--5 days} and {96.5\%} of safety violations eliminated.
Our results show that LLMs, augmented with domain-specific constraints and personalized memory, can effectively manage complex multi-parameter system optimization tasks on mobile devices.


\bibliographystyle{ACM-Reference-Format}
\bibliography{ref}

\end{document}